\documentclass[letterpaper]{article} 
\usepackage{aaai24}  
\usepackage{times}  
\usepackage{helvet}  
\usepackage{courier}  
\usepackage[hyphens]{url}  
\usepackage{graphicx} 
\usepackage{natbib}  
\usepackage{caption} 
\usepackage{algorithm}
\usepackage{algorithmic}

\usepackage{newfloat}
\usepackage{listings}
\DeclareCaptionStyle{ruled}{labelfont=normalfont,labelsep=colon,strut=off} 
\floatstyle{ruled}
\newfloat{listing}{tb}{lst}{}
\floatname{listing}{Listing}
\title{Structurally guided task decomposition in spatial navigation tasks}

\title{Structurally guided task decomposition in spatial navigation tasks}
\author {
    Ruiqi He\textsuperscript{\rm 1},
    Carlos G. Correa \textsuperscript{\rm 2},
    Thomas L. Griffiths \textsuperscript{\rm 3,4},
    Mark K. Ho\textsuperscript{\rm 5}
}
\affiliations {
    \textsuperscript{\rm 1}Max Planck Institute for Intelligent Systems, Tübingen, Germany\\
    \textsuperscript{\rm 2} Princeton Neuroscience Institute, Princeton University, Princeton, New Jersey, United States of America\\
    \textsuperscript{\rm 3} Department of Psychology, Princeton University, Princeton, New Jersey, United States of America\\
    \textsuperscript{\rm 4} Department of Computer Science, Princeton University, Princeton, New Jersey, United States of America\\
    \textsuperscript{\rm 5} Department of Computer Science, Stevens Institute of Technology, Hoboken, New Jersey, United States of America\\
    ruiqi.he@tuebingen.mpg.de, Max-Planck-Ring 4, 72076 Tübingen, +49 7071 601-1700
}

\begin{document}

\maketitle

\begin{abstract}
How are people able to plan so efficiently despite limited cognitive resources? We aimed to answer this question by extending an existing model of human task decomposition that can explain a wide range of simple planning problems by adding structure information to the task to facilitate planning in more complex tasks. The extended model was then applied to a more complex planning domain of spatial navigation. Our results suggest that our framework can correctly predict the navigation strategies of the majority of the participants in an online experiment. 
\end{abstract}

\section{Introduction}
People plan hierarchically\textemdash from planning the next holiday to their long-term career. Previous studies have investigated the principles that guide how people structure their behavioral hierarchies based on how they can simplify representational and computational costs \citep{solway2014,correa2023humans} in simple and unstructured planning domains. We build on a task decomposition framework introduced by \citet{correa2023humans}, which was able to predict hierarchical behavior in people in graph-structured planning tasks. 
While the framework accounts for the complexity of planning, it does not explain how people can plan so efficiently despite limited cognitive resources. One explanation could be the incorporation of structural information about the task to facilitate planning. \citet{binder2021visual} suggest that people exploit visual structure to inform their planning. We thus extended the framework of task decomposition to include structured information about the planning task through a heuristic search based on spatial distance and applied it to a more complex spatial navigation task. We tested our novel framework on a navigation planning experiment that consists of a large family of tasks, where participants must choose between two paths to take in order to solve a maze (see Figure \ref{fig:mazeexample}). Our results suggest that our framework can predict the navigation choice of the majority of the participants in an online experiment.

\section{Methodology}
Our modeling approach builds on the framework of task decomposition by minimizing the computational cost of planning while maximizing task utility \citep{correa2023humans}. The framework consists of three levels: 

\paragraph{Task decomposition} decomposes a task into subtasks such that overall computational costs are minimized.
We formalize the task as $(\mathcal{S}, T, s_0, g)$, which is defined by the set of possible states, $\mathcal{S}$; an initial state, $s_0$; a goal state, $g$; and the set of possible state transitions $T \subseteq \mathcal{S} \times \mathcal{S}$, so that $s$ can transition to $s'$ when $(s, s') \in T$.
Each subtask $z$ is simply defined as reaching a subgoal state $z_{sg}$\footnote{Note that the subtask is essentially an option \citep{sutton1999between} that terminates at the subgoal, making subtask-level planning a semi-Markov decision process.}. 
In our setting, we assume the task has been decomposed into a fixed set of subtasks $\mathcal{Z}$ for simplicity, consisting of
the two states adjacent to $s_0$, a simple approach for formalizing the choice between the two halves of the maze.

\paragraph{Subtask-level planning} decides which subtask to choose based on the expected reward and computational cost of visiting it en route to the goal. The 
subtask that maximizes this overall reward is
\begin{equation}
    z^* = \arg \max_{z \in \mathcal{Z}} R_{\texttt{Alg}}(s_0, z) + R_{\texttt{Alg}}(z, g).
    \label{eq:subtask}
\end{equation}
We assume the overall task of reaching $g$ can also be a subtask for the purpose of action-level planning. Note that while we fixed subgoals above, the choice between subtasks here still balances task reward and computational cost.

\paragraph{Action-level planning} which finds a sequence of actions to accomplish a subtask. 
Action-level planning finds a sequence of states from a start state $s$ until the subtask $z$ is completed by reaching $z_{sg}$, $\pi = \langle s_0, s_1, ..., z_{sg} \rangle$.
Shorter plans are preferred, formalized by the reward function $R(\pi)=-|{\pi}|$.
A planning algorithm $\texttt{Alg}$ non-deterministically returns a plan and algorithm run-time, $P_{\texttt{Alg}}(\pi, t \mid s, z)$.
So the expected reward for reaching the subtask $z$, while accounting for algorithm run-time, is
$R_{\texttt{Alg}}(s, z) = \sum_{\pi, t}P_{\texttt{Alg}}( \pi, t \mid s, z)  \left[R(\pi) - t  \right]$.


We incorporate structural information by using A* \citep{hart1968formal}  with a spatial heuristic cost based on the Manhattan distance defined as $h(s; g) = | s_x - g_x | + | s_y - g_y |$ given states with coordinates $s=(s_x, s_y)$. The algorithm run-time is computed by counting the number of visited states during search.


\section{Experimental design and results}
\begin{figure}
    \centering
    \includegraphics[width=0.6\linewidth]{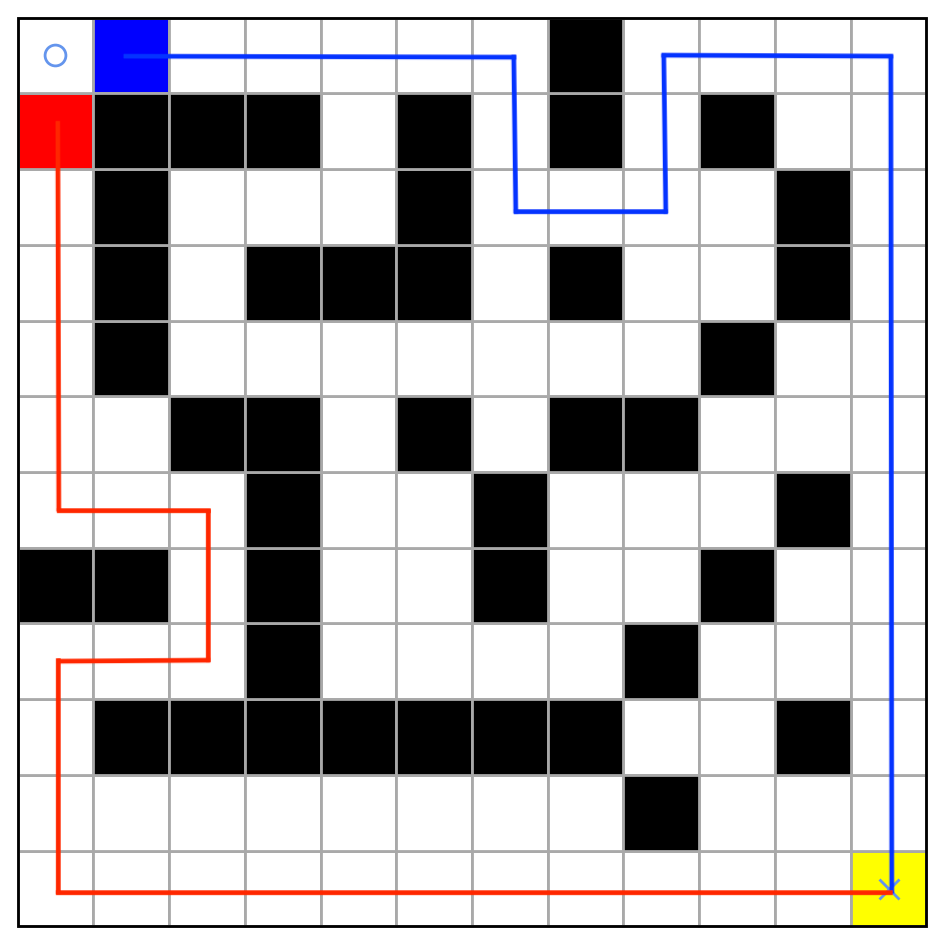}
    \caption{One example of the maze, where the participants were asked to navigate from the blue dot to the goal (yellow tile in the lower right corner). The red and blue subgoals were not visible to the participants but marked the two subtasks. The subtask associated with the red path has a lower planning cost (planning cost 53, step cost 26, total cost: 79) than the other subtask with the blue path (planning cost 63, step cost 26, total cost 89). In addition to both subtasks having the same optimal path mirrored along the diagonal, the number of walls in each triangle is also the same.}
    \label{fig:mazeexample}
\end{figure}

To test our hypothesis, we designed 6 base mazes and created 8 transformations through rotations and flipping along diagonal, horizontal, and vertical axes resulting in a total of 6 sets, each containing 8 mazes. Each of the resulting 48 mazes can be decomposed into two subtasks, where one subtask was designed to have a higher planning cost than the other subtask. The optimal route for both subtasks is the same but mirrored along the diagonal (see Figure \ref{fig:mazeexample}). Our model hypothesizes that people would choose the subtask with the lower overall cost, which is the sum of the optimal path length and planning cost. 

To test this hypothesis, we recruited 41 participants (one participant was excluded due to not completing the task) on Prolific to each navigate through 12 different mazes (two mazes were randomly sampled from each set to ensure that participants saw two of each base maze). Before playing the navigation game, participants received instructions and played two practice trials to familiarize themselves with the task. Participants were given up to one minute in the beginning for planning, after which they were motivated to reach the goal as quickly as possible through a performance-dependent bonus that rewarded shorter paths. 

\begin{figure}
    \centering
    \includegraphics[width=0.7\linewidth, height=120px]{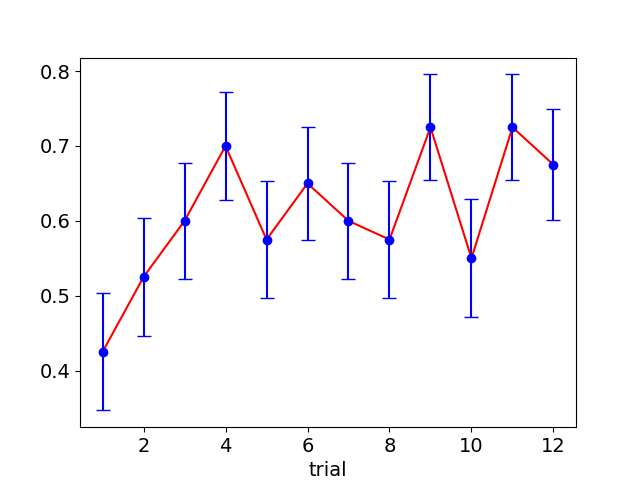}
    \caption{Proportion of the favorable subtask chosen and error bars showing the standard error at each trial}
    \label{fig:proportions}
\end{figure}

We observed that after the third trial, the majority of participants chose the subtask which, according to our model, is more favorable due to a lower planning cost (see Figure \ref{fig:proportions}). Modeling the path choice in the last trial using logistic regression ($path\_chosen \sim intercept$), where the more favorable path was encoded as 1, resulted in an intercept of $0.731$ ($sd=0.338, p=0.030$). 
In addition, we observed a learning process among the participants. The proportion of the computationally simpler subtask increased from 42.5\% in the first trial to 67.5\% in the last trial. This increasing trend was confirmed by again fitting a logistic regression model ($path\_chosen \sim intercept + trial\_number$) resulting in a significant positive slope with a coefficient of $0.063$ ($sd = 0.027, p=0.021$) and $0.106$ intercept ($sd = 0.174, p=0.542$).

\section{Discussion and future work}
In this work, we presented an extension to the task decomposition framework of \citet{correa2023humans} and tested it on a more complex planning domain of spatial navigation tasks. Our model can capture the navigation strategies of the majority of the participants.  However, we see that after 12 trials, 32.5\% of the participants did not choose the subtask with lower planning costs. This suggests room for improvement in the choice of planner and/or heuristics. 
In addition, we observe a learning process. Future work can therefore concentrate on how people \textit{learn} to decompose and select tasks. 

This line of research elegantly bridges our understanding of the human mind and comprehending human planning processes with the development of efficient algorithms for advanced machine learning systems.

\bibliography{aaai24}

\end{document}